\pdfoutput=1
\documentclass[11pt]{article}

\usepackage{acl}

\usepackage{times}
\usepackage{latexsym}
\usepackage{booktabs}
\usepackage{microtype}
\usepackage{amssymb}
\usepackage{pifont}
\usepackage{xcolor} % Include this in the preamble
\usepackage{graphicx}
\usepackage{placeins}
\usepackage{natbib}
\usepackage{tablefootnote}
\usepackage{longtable}
\usepackage{cleveref}
\usepackage{enumitem}
\usepackage{tikz}
\usepackage{booktabs}
\usepackage{forest}
\usetikzlibrary{trees}
\usepackage{pdfpages}
\usepackage{enumitem}

\usepackage[T1]{fontenc}
\usepackage{tabularx}
\usepackage{multirow}
\usepackage{bigfoot}
\DeclareNewFootnote{ANote}[fnsymbol]

\usepackage[textsize=footnotesize]{todonotes}
\usepackage{linguex}
\usepackage{subcaption}
\usepackage{color, colortbl}
\newcommand{\best}{\cellcolor{blue!30}}

\usepackage{xspace}

\alignSubExtrue

\usepackage[verbose]{newunicodechar}

\usepackage{inconsolata}

\definecolor{Gray}{gray}{0.9}

\title{SemEval-2024 Task 1: Semantic Textual Relatedness\\ for African and Asian Languages}

\author{Nedjma Ousidhoum$^{1}$\thanks{$^{*}$Equal contribution from first and second authors, authors 3 to 16 are alphabetically ordered.} , Shamsuddeen Hassan Muhammad$^{2*}$ , Mohamed Abdalla, \\
{\bf Idris Abdulmumin$^3$, Ibrahim Said Ahmad$^{4}$, Sanchit Ahuja$^{5}$, Alham Fikri Aji$^6$,
} \\
{\bf Vladimir Araujo$^{7}$, Meriem Beloucif$^{8}$,
  Christine De Kock$^{9}$, Oumaima Hourrane,
}\\
{\bf Manish Shrivastava$^{10}$, Thamar Solorio$^{6}$, Nirmal Surange$^{10}$, Krishnapriya Vishnubhotla$^{11}$,}\\
{\bf Seid Muhie Yimam$^{12}$, Saif M. Mohammad$^{13}$}
\\
 \footnotesize {$^1$Cardiff University, $^{2}$Imperial College London, $^{3}$Data Science for Social Impact Research Group, 
 University of Pretoria,} \\
 \footnotesize {$^4$Institute For Experiential AI,
 Northeastern University, $^5$BITS Pilani, $^6$MBZUAI, $^{7}$KU Leuven, $^{8}$Uppsala University,} \\
  \footnotesize {$^{9}$The University of Merlbourne, $^{10}$IIIT Hyderabad, $^{11}$University of Toronto, $^{12}$Universität Hamburg,}\\
 \footnotesize \texttt{$^{13}$National Research Council Canada}. \\
 \footnotesize \texttt{Contact: OusidhoumN@cardiff.ac.uk, s.muhammad@imperial.ac.uk}
 }
 
\begin{document}

\maketitle

\begin{abstract}
We present the first shared task on Semantic Textual Relatedness (STR). While earlier shared tasks primarily focused on semantic similarity, we instead investigate the broader phenomenon of semantic relatedness across 14 languages: \textit{Afrikaans, Algerian Arabic, Amharic, English, Hausa, Hindi, Indonesian, Kinyarwanda, Marathi, Moroccan Arabic, Modern Standard Arabic, Punjabi, Spanish,} and \textit{Telugu}. These languages originate from five distinct language families and are predominantly spoken in Africa and Asia -- regions characterised by the relatively limited availability of NLP resources. Each instance in the datasets is a sentence pair associated with a score that represents the degree of semantic textual relatedness between the two sentences. Participating systems were asked to rank sentence pairs by their closeness in meaning (i.e., their degree of semantic relatedness) in the 14 languages in three main tracks: (a)\ supervised, (b)\ unsupervised, and (c)\ crosslingual. 
The task attracted 163 participants. We received 70 submissions in total (across all tasks) from 51 different teams, and 38 system description papers. 
We report on the best-performing systems as well as the most common and the most effective approaches for the three different tracks.

\end{abstract}

\section{Introduction}
Defining the relationship between two units of text is an important component of constructing text representations. Within this context, semantic textual relatedness (STR) aims to capture the degree to which two linguistic units (e.g., words or sentences, etc.) are close in meaning \cite{mohammad2012distributional}. Two units may be related in a variety of different ways (e.g., by expressing the same view, originating from the same time period, elaborating on each other, etc.). On the other hand, semantic textual similarity (STS) considers only a narrow view of the relationship that may exist between texts (such as equivalence or paraphrase) which does not incorporate other dimensions of relatedness such as entailment, topic or view similarity, or temporal relations \cite{abdalla-etal-2023-makes}. For example, \textit{`I am feeling sick.'} and \textit{`Get well soon!'} would receive a low similarity score, despite the two being very related. In this shared task, we investigate the broader concept of semantic textual relatedness. 
STR is central to understanding meaning in text \cite{halliday1976cohesion,miller1991contextual,morris1991lexical} and its automation can benefit various downstream tasks such as evaluating sentence representation methods, question answering, and summarisation \cite{abdalla-etal-2023-makes,wang-etal-2022-just}.

Prior shared tasks \cite{agirre2012semeval,agirre2013sem,agirre2014semeval,agirre2015semeval,agirre2016semeval,cer-etal-2017-semeval} have mainly focused on textual similarity. In this work, we provide participants with SemRel \cite{ousidhoum2024semrel2024}, a collection of 14 newly curated monolingual STR datasets for Afrikaans \texttt{(afr)}, Amharic \texttt{(amh)}, Modern Standard Arabic \texttt{(arb)}, Algerian Arabic \texttt{(arq)}, Moroccan Arabic \texttt{(ary)}, English \texttt{(eng)}, Spanish \texttt{(esp)}, Hausa \texttt{(hau)}, Hindi \texttt{(hin)}, Indonesian \texttt{(ind)}, Kinyarwanda \texttt{(kin)}, Marathi \texttt{(mar)}, Punjabi \texttt{(pun)} and Telugu \texttt{(tel)}. 
The datasets are composed of sentence pairs, each assigned a relatedness score between 0 (completely unrelated) and 1 (maximally related) with a large range of expected relatedness values. The pairs of sentences were first selected from pre-existing datasets covering various topics and formality levels, e.g., news data, Wikipedia, and conversational data. To generate the relatedness scores, the sentence pairs were then annotated by native speakers who performed comparisons between different pairs of sentences using Best--Worst Scaling (BWS) \cite{louviere1991best,kiritchenko2017best}. The shared task included three main tracks: (1)\ supervised, (2)\ unsupervised, and (3)\ cross-lingual. 

 Each team could provide submissions for one, two, or all of the tracks in one or more languages. Our official evaluation metric was the Spearman rank correlation coefficient, which captures how well the system-predicted rankings of test instances aligned with human judgments. Our task attracted 163 participants, received 70 final submissions from 51 different teams, and 38 teams submitted system description papers. Track A (supervised) received the largest number of submissions: 40, followed by 18 submissions for track B (unsupervised) and 12 for track C (crosslingual). Most teams participated in multiple languages (more than eight on average). All of the task details and resources are available on the task website.\footnote{\href{https://semantic-textual-relatedness.github.io}{https://semantic-textual-relatedness.github.io}}

\begin{table}[t]
\small
\centering
    % \begin{tabular}{lllll}
       \begin{tabular}{llrrr}
    \toprule[1.2pt]
        \textbf{Lang.} & \textbf{Family} & \textbf{Train} & \textbf{Dev} & \textbf{Test} \\ \midrule
        \texttt{afr} & Indo-European & - & 375 & 375\\
        \texttt{amh} & Afro-Asiatic & 992 & 95 & 171 \\
        \texttt{arb} & Afro-Asiatic & - & 32 & 595 \\
        \texttt{arq} & Afro-Asiatic & 1,261 & 97 & 583 \\
        \texttt{ary} & Afro-Asiatic & 925 & 70 & 427 \\
        \texttt{eng} & Indo-European & 5,500 & 250 & 2,600 \\
        \texttt{esp} & Indo-European & 1,562 & 140 &  600\\
        \texttt{hau} & Afro-Asiatic & 1,763 & 212 & 603 \\
        \texttt{hin} & Indo-European & - & 288 & 968 \\
        \texttt{ind} & Austronesian & - & 144 & 360 \\
        \texttt{kin} & Niger-Congo & 778 & 102 & 222 \\
        \texttt{mar} & Indo-European & 1,200 & 293 & 298\\
        \texttt{pan} &  Indo-European & - & 638& 242 \\
        \texttt{tel} & Dravidian & 1,170 & 130 & 297\\ \bottomrule
    \end{tabular}
\caption{The language families and data split sizes of the different datasets. Datasets with no traning sets were only used in tracks B and C.}
\label{tab:data_collection}
\end{table}

\section{Related Work}
\label{sec:related}

 The field of semantic textual relatedness in natural language processing covers a variety of approaches and techniques designed to measure the closeness in meaning between units of text, specifically words \cite{miller-1994-wordnet} or sentences \cite{abdalla-etal-2023-makes}. 
 
 Most prior shared tasks focus on semantic textual similarity, a narrower subset of relatedness, and often only cover high-resource languages such as English \cite{agirre2012semeval,agirre2013sem,agirre2014semeval,agirre2015semeval,agirre2016semeval}, Arabic, German, Spanish, and Turkish \cite{cer-etal-2017-semeval} with few exceptions such as \citet{armendariz-etal-2020-semeval} who also included Slovene, Finnish, and Croatian.
 
By comparison, this shared task focuses on sentence-level STR in various low-resource languages. To our knowledge, the only corpora specially designed for semantic textual relatedness between pairs of sentences was created by \citet{abdalla-etal-2023-makes} for English.
The core of \citet{abdalla-etal-2023-makes} approach served as the model for data annotations added to new ways of data collection--curation for several less-resourced languages.

\section{Data}\label{sec:data}
 \begin{table*}[]
    \centering
    \resizebox{\textwidth}{!}{
    \begin{tabular}{c|cccccccccccccc}
         \toprule
         \textbf{Language} & \textbf{\texttt{afr}} & \textbf{\texttt{amh}} & \textbf{\texttt{arb}} & \textbf{\texttt{arq}} & \textbf{\texttt{ary}} & \textbf{\texttt{eng}} & \textbf{\texttt{esp}} & \textbf{\texttt{hau}} & \textbf{\texttt{hin}} & \textbf{\texttt{ind}} & \textbf{\texttt{kin}} & \textbf{\texttt{mar}} & \textbf{\texttt{pun}} & \textbf{\texttt{tel}}  \\ 
    \midrule
             \textbf{\#Annotators} & 2 & 4 & 2-3 & 2 & 2 & 2-4 & 2-4 & 2-4 & 4 & 2 & 2 & 2-3 & 2 & 4 \\ \midrule
        \textbf{SHR train/dev} & 0.85 & 0.89 & 0.86 & 0.64 & 0.77 & 0.80 & 0.70 & 0.74 & 0.93 & 0.68 & 0.74 & 0.92 & 0.65 & 0.79 \\ \midrule
        \textbf{SHR test} & 0.85 & 0.89 & 0.86 & 0.64 & 0.77 & 0.80 & 0.70 & 0.74 & 0.93 & 0.68 & 0.74 & 0.92 & 0.65 & 0.79 \\

    \bottomrule
    \end{tabular}
    }
    \caption{SHR (split-half reliability) scores for each of the created dataset splits and numbers of annotators per tuple (\#Annotators).}
    \label{tab:shr_scores}
\end{table*}
\label{subsec:annotation}

\subsection{Data Collection}
\label{subsec:collection}
A key step in the data creation process was identifying text sources for each language and selecting sentence pairs. This was particularly challenging for low-resource languages such as Hausa, Telugu, or Algerian Arabic. 
Since most SemRel languages are low-resource, the domain, (in)formality, and diversity of the sentence pairs were highly dependent on the publicly available corpora.
We aimed to collect datasets with average-length sentences, free of offensive utterances, and as diverse as possible. Thus, data instances were extracted for each language using a tailored combination of heuristics such as lexical overlap and paraphrases. 
We used further pre-processing, post-processing, and data analysis methods to avoid incoherence and unnaturalness. 

Since arbitrarily selecting sentences and pairing them would lead to many unrelated instances, we relied on the following heuristics to pair sentences and ensure that the pairs would exhibit relatedness scores varying from completely unrelated to very related:
\begin{enumerate}[noitemsep,nolistsep]
    \item \textbf{Lexical Overlap} Select sentences with various proportions of lexical overlap, i.e., one or more words/tokens in common, with or without using TF/IDF normalisation. 
    \item \textbf{Contiguity/Entailment} Select adjacent pairs of sentences in a paragraph or a social media thread, i.e., sentences that appear one after the other.
    \item \textbf{Paraphrases or Machine Translation (MT) Paraphrases} Select pairs of sentences from paraphrase or MT data. For MT, we pivot across the translation and back to the source language to generate a new sentence and pair it with the original. 
    \item \textbf{Random selection} Random pairs of sentences are selected.
    \vspace*{-1mm}
\end{enumerate}

We elaborate on the detailed data collection and processing steps in \citet{ousidhoum2024semrel2024}.

\subsection{Data Annotation}
 As the notions of \textit{related} and \textit{unrelated} do not have clear boundaries with no unanimous definition in the literature, we use comparative annotations and rely on the intuitions of fluent speakers for each language to choose between sentence pairs. Therefore, instead of relying on vague class definitions, we capture common perceptions of semantic relatedness (i.e., what is believed by the vast majority) rather than ``correct'' or ``right'' rankings.

We used Best--Worst Scaling (BWS) \cite{louviere1991best,kiritchenko2017best}, a form of comparative annotation that avoids various biases of traditional rating scales, to annotate our data instances and generate an ordinal ranking of instances.
In BWS, annotators are given $n$ items (an $n$-tuple, where $n >1 $ and commonly $n = 4$). They are asked which item is the {\it best} (highest in terms of the property of interest) and which is the {\it worst} (lowest in terms of the property of interest).
When working on $4$-tuples, best--worst annotations are particularly efficient because each best and worst annotation will reveal the order of five of the six-item pairs. 
Real-valued scores of association 
between the items and the property of interest can be determined using simple arithmetic on the number of times an item was chosen best and the number of times it was chosen worst \cite{Orme_2009,flynn2014}.  
It has been empirically shown that annotations for $2N$ $4$-tuples are sufficient for obtaining reliable scores (where $N$ is the number of items) \cite{louviere1991best,maxdiff-naacl2016}. 
Kiritchenko and Mohammad \shortcite{KiritchenkoM2017bwsvsrs} showed through empirical experiments that BWS produces more reliable and discriminating scores than those obtained using rating scales. (See \cite{maxdiff-naacl2016,KiritchenkoM2017bwsvsrs} for further details on BWS.)
We generated tuples using the BWS scripts provided by \citet{kiritchenko2017best}\footnote{\href{https://saifmohammad.com/WebPages/BestWorst.html}{https://saifmohammad.com/WebPages/BestWorst.html}}.

We report the number of annotators and the split-half reliability (SHR) scores \cite{cronbach1951coefficient,kuder1937theory} for each of the datasets in Table \ref{tab:shr_scores}. 
SHR measures the degree to which repeating the annotations results in similar relative rankings of the instancesOverall the scores in Table \ref{tab:shr_scores} vary between 0.64 and 0.96, which indicates a high annotation reliability.

\section{Task Description}
\definecolor{Gray}{gray}{0.9}

\begin{table*}[]
    \centering
    % \resizebox{\textwidth}{!}{
{\small
    \begin{tabular}{ccrccrccr}
    \toprule
        & \multicolumn{3}{c}{\textbf{Track A (Supervised)}} & \multicolumn{3}{c}{\textbf{Track B (Unsupervised)}} & \multicolumn{2}{c}{\textbf{Track C (Crosslingual)}} \\ \cmidrule(lr){2-3}\cmidrule(lr){4-6}\cmidrule(lr){7-9}
        \textbf{\#} & \textbf{Team} & \textbf{Score} & & \textbf{Team} & \textbf{Score} 
        & & \textbf{Team} & \textbf{Score} \\
    \midrule
    \rowcolor{Gray} &  &
     & * & Lexical Overlap & 0.456    & & &  \\
    \midrule
    \rowcolor{Gray}
    * & baseline (LaBSE)& 0.762 & * & baseline (XLMR)& 0.353  & * & baseline (LaBSE) & 0.579 \\
    \midrule
        \textbf{1} & AAdam & 0.800 & &SATLab & 0.543 &  & AAdaM & 0.650 \\
        \textbf{2} & NRK & 0.781 & & MasonTigers & 0.514 & &  UAlberta & 0.589 \\
        %\textbf{2} & king001 & 0.812 & MasonTigers & 0.514 &  king001 & 0.641\\
        \textbf{3} & PEAR & 0.758 & & HW--TSC & 0.482 & & silp\_nlp & 0.566\\

        \textbf{4} & silp\_nlp & 0.740 & & UAlberta & 0.481 & &  MaiNLP & 0.499 \\
        %USTC\_NLP & 0.523 \\
        
        \textbf{5} & NLP\_1@SSN & 0.740 & & silp\_nlp & 0.400 & & ustcctsu & 0.445 \\
        %\textbf{11} & DN & 64.88 & & & & \\
    \bottomrule
    \end{tabular}
    
    \caption{Top 5 submissions per track. See Appendix for paper information about the different teams.  \textbf{*} shows baseline results using lexical overlap, XLMR and LaBSE reported in the SemRel dataset paper \cite{ousidhoum2024semrel2024}.}
    \label{tab:top_systems}
    }
\end{table*}

%\subsection{Task Definition}

In this task, we aim to predict the semantic textual relatedness (STR) of sentence pairs. Participants had to rank sentence pairs by their degree of semantic relatedness which varies between 0 (unrelated) and 1 (closely related). Each team could provide submissions for one, two, or all of the tracks presented below.

\subsection{Track A: Supervised}

Participants were to submit systems trained on the labeled training datasets provided. Participating teams were allowed to use any publicly available datasets (e.g., other relatedness and similarity datasets or datasets in any other languages). However, they had to report on additional data they used, and ideally report how each resource impacted the final results.

\subsection{Track B: Unsupervised}

Participants were to submit systems that were developed without the use of any labeled datasets pertaining to semantic relatedness or semantic similarity between units of text more than two words long in any language. The use of unigram or bigram relatedness datasets (from any language) was permitted.

\subsection{Track C: Cross-lingual}

Participants were to submit systems that were developed without the use of any labeled semantic similarity or semantic relatedness datasets in the target language and with the use of labeled dataset(s) from at least one other language. Using labeled data from another track was mandatory for a submission to this track.

\subsection{Official Evaluation Metric}
The official evaluation metric for this task is the Spearman rank correlation coefficient, which captures how well the system-predicted rankings of test instances align with human judgments. We provided the participants with an evaluation script on GitHub page\footnote{\href{https://github.com/semantic-textual-relatedness/Semantic_Relatedness_SemEval2024}{https://github.com/semantic-textual-relatedness/Semantic\_Relatedness\_SemEval2024}}.

\subsection{Task Organisation}
We released some pilot datasets before the start of the shared task for participants to have a better understanding of the task (i.e., the datasets, the languages involved, and the labels) and provided the participants with a starter kit on GitHub.%#shared-task-starter-kit}}. 

\section{Evaluation}

\subsection{Our baselines}
In \Cref{tab:top_systems}, we report a simple lexical overlap baseline which consists of the Dice coefficient between two sentences A and B: the number of unique unigrams occurring in both sentences, adjusted by their lengths \cite{abdalla-etal-2023-makes}: 
% It is defined as follows:
\begin{equation}
        \frac{2 \times | unigram(A) \cap unigram(B) |}{ | unigram(A) + unigram(B) |}
\end{equation}

In addition, we used LaBSE (Label Agnostic BERT Sentence Embeddings) \cite{feng2020language} which can map 109 languages into a shared vector space. With the embeddings covering all the SemRel languages, we report baseline results using the default hyperparameters set in the sentence-transformers repository\footnote{\href{https://github.com/UKPLab/sentence-transformers}{https://github.com/UKPLab/sentence-transformers}}. We used:
\begin{itemize}[noitemsep,nolistsep] 
    \item the predefined setup without further fine-tuning,
    \item the LaBSE model further fine-tuned on our training data using a cosine similarity loss.
\end{itemize}
%\paragraph{Crosslingual settings}
For the crosslingual baselines, we fine-tuned LaBSE on the English training set and tested on all the other datasets except English while using the Spanish training set to fine-tune LaBSE when testing on English. We elaborate on
the detailed baseline experiment in \cite{ousidhoum2024semrel2024}

\subsection{Participating Systems and Results}

\subsection{Participant Overview}

During the evaluation phase, 163 people registered for the competition. Of these, 51 teams made 70 final submissions across tracks \footnote{The details can be found in the Appendix.}. Track A received 40 final submissions, track B received 12 submissions, and track C received 18. For track A, most participants submitted systems for at least eight languages. We report the top--5 performing systems in all tracks in Table \ref{tab:top_systems}.

\subsection{Task A: Supervised}
\begin{table*}[!ht]
    \centering
    \resizebox{\textwidth}{!}{%
    \begin{tabular}{llcccccccccc}
    \toprule
        \textbf{Rank} & \textbf{Team} &  \textbf{\texttt{amh}}  & \textbf{\texttt{arq}} & \textbf{\texttt{ary}} & \textbf{\texttt{eng}}& \textbf{\texttt{esp}} & \textbf{\texttt{hau}} & \textbf{\texttt{kin}} & \textbf{\texttt{mar}} & \textbf{\texttt{tel}} & \textbf{\texttt{Average}} \\

    \midrule

    \textbf{1} & AAdaM \cite{zhang-EtAl:2024:SemEval20242} & 0.867 & 0.662 & 0.835 & 0.848 & \textbf{0.740} & 0.724 & 0.779 & 0.894 & 0.848 & 0.800 \\
    \textbf{2} & NRK  \cite{nguyen-thin:2024:SemEval2024} & 0.864 & 0.674 & 0.827 & 0.833 & 0.690 & 0.672 & 0.757 & 0.879 & 0.834 & 0.781 \\
    
    \midrule
          \rowcolor{Gray}
    * & SemRel baseline (LaBSE) & 0.789 & 0.847 & 0.761 & 0.830 & 0.702 & 0.693 & 0.725 & 0.881 & 0.817 &  0.762  \\
      \midrule
      
    \textbf{3} & PEAR \cite{jrgensen:2024:SemEval2024} & 0.834 & 0.463 & 0.815 & 0.848 & 0.710 & 0.694 & 0.772 & 0.856 & 0.827 & 0.758 \\

    \textbf{4} & silp\_nlp \cite{singh-goyal-tiwary:2024:SemEval2024} & 0.837 & 0.594 & 0.808 & 0.845 & 0.658 & 0.724 & 0.485 & 0.863 & 0.843 & 0.740 \\
    \textbf{5} & NLP\_1@SSN \cite{b-EtAl:2024:SemEval2024}& - & 0.623 & 0.745 & 0.835 & 0.705 & 0.628 & 0.723 & 0.871 & 0.789 & 0.740 \\
    \textbf{6} & UAlberta \cite{shi-EtAl:2024:SemEval2024} & 0.854 & 0.464 & 0.497 & 0.853 & 0.705 & 0.735 & 0.641 & 0.890 & 0.857 & 0.722 \\
    \textbf{7} & MBZUAI-UNAM \cite{ortizbarajas-belenguix-gomzadorno:2024:SemEval2024} & 0.840 & 0.541 & 0.786 & 0.832 & 0.697 & 0.670 & 0.458 & 0.867 & 0.785 & 0.720 \\
    \textbf{8} & INGEOTEC  \cite{moctezuma-tellez-graff:2024:SemEval2024} & 0.702 & 0.566 & 0.811 & 0.809 & 0.678 & 0.576 & 0.630 & 0.784 & 0.801 & 0.706 \\
    \textbf{9} & HausaNLP \cite{salahudeen-EtAl:2024:SemEval2024} & 0.353 & 0.587 & 0.834 & 0.794 & 0.723 & 0.594 & 0.633 & 0.837 & 0.800 & 0.684 \\
    \textbf{10} & KINLP & - & 0.471 & 0.779 & 0.740 & 0.581 & 0.616 & 0.763 & 0.749 & 0.754 & 0.682 \\
    \textbf{11} & BITS Pilani \cite{venkatesh-raman:2024:SemEval2024} & 0.800 & 0.510 & 0.444 & 0.832 & 0.656 & 0.508 & 0.518 & 0.842 & 0.814 & 0.658 \\
    \textbf{12} & OZemi  \cite{takahashi-EtAl:2024:SemEval2024} & 0.781 & 0.371 & 0.445 & 0.805 & 0.620 & 0.620 & 0.567 & 0.862 & 0.782 & 0.650 \\
    \textbf{13} & Text Mining \cite{keinan:2024:SemEval2024} & 0.713 & 0.443 & 0.701 & 0.720 & 0.661 & 0.543 & 0.413 & 0.778 & 0.706 & 0.631 \\
    \textbf{14} & MasonTigers \cite{goswami-EtAl:2024:SemEval2024} & 0.785 & 0.400 & 0.376 & 0.836 & 0.651 & 0.477 & 0.367 & 0.818 & 0.802 & 0.612 \\
    \textbf{15} & YSP \cite{aali-hamidian-farinneya:2024:SemEval2024} & 0.643 & 0.402 & - & 0.819 & 0.635 & 0.387 & 0.315 & 0.689 & 0.643 & 0.567 \\
    %\midrule
    %\rowcolor{Gray}
    %* & Lexical Overlap baseline & 0.633 & 0.400 &	0.627 &	0.670 &	0.670 & 0.306 &	0.333 &	0.619 &	0.697 &	0.550  \\
     % \midrule
    \textbf{16} & IITK \cite{basak-EtAl:2024:SemEval2024} & 0.550 & 0.339 & 0.358 & 0.808 & 0.591 & 0.219 & 0.138 & 0.666 & 0.282 & 0.439 \\
    \textbf{17} & YNUNLP2023 \cite{li-wang-zhang:2024:SemEval2024} & 0.789 & 0.235 & 0.092 & 0.557 & 0.404 & 0.269 & 0.186 & 0.544 & 0.617 & 0.410 \\
    \textbf{NR} & PALI & \best \textbf{0.889} & 0.679 & \best \textbf{0.863} & \best \textbf{0.860} & 0.724 & \best \textbf{0.764} & 0.813 & \best \textbf{0.911} & 0.864 & \best \textbf{0.819}\\
    \textbf{NR} & king001 & 0.888 & \best \textbf{0.682} & 0.860 & 0.843 & 0.721 & 0.747 & \best \textbf{0.817} & 0.897 & 0.853 & 0.812 \\
    \textbf{NR} & saturn & 0.845 & 0.578 & 0.798 & - & - & 0.699 & 0.755 & 0.873 & \best \textbf{0.873} & 0.774 \\
    \textbf{NR} & UMBCLU \cite{roydipta-vallurupalli:2024:SemEval2024} & - & - & 0.745 & 0.838 & 0.721 & 0.640 & 0.681 & 0.841 & 0.682 & 0.733 \\ 
    \textbf{NR} & SemanticCUETSync \cite{hossain-EtAl:2024:SemEval2024} & - & - & - & 0.822 & 0.677 & - & - & 0.870 & 0.820 & 0.796 \\
    \textbf{NR} & NLP-LISAC \cite{benlahbib-EtAl:2024:SemEval2024} & - & 0.604 & 0.789 & 0.835 & 0.717 & - & - & - & - & 0.736 \\
    \textbf{NR} & Unknown & - & - & - & 0.831 & - & - & - & 0.882 & 0.841 & 0.852 \\
    \textbf{NR} & BpHigh & - & - & - & 0.809 & - & - & - & 0.875 & 0.769 & 0.819 \\
    \textbf{NR} & Sharif\_STR \cite{ebrahimi-EtAl:2024:SemEval20242} & - & 0.380 & - & 0.827 & 0.673 & - & - & - & - & 0.441 \\
    \textbf{NR} & CAILMD-23 \cite{sonavane-EtAl:2024:SemEval2024} & - & - & - & 0.823 & - & - & - & 0.871 & - & 0.847 \\
    \textbf{NR} & WarwickNLP  \cite{ebrahim-joy:2024:SemEval2024} & - & - & 0.816 & 0.842 & - & - & - & - & - & 0.829 \\
    \textbf{NR} & GIL-IIMAS UNAM & - & - & - & 0.830 & 0.731 & - & - & - & - & 0.780 \\
    \textbf{NR} & msiino & - & - & - & 0.809 & 0.611 & - & - & - & - & 0.710 \\
    \textbf{NR} & NLU-STR \cite{malaysha-jarrar-khalilia:2024:SemEval2024} & - & 0.525 & 0.832 & - & - & - & - & - & - & 0.678 \\
    \textbf{NR} & Tübingen-CL \cite{zhang-ltekin:2024:SemEval2024} & - & - & - & 0.850 & - & - & - & - & - & 0.850 \\
    \textbf{NR} & Pinealai \cite{eponon-ramosperez:2024:SemEval2024} & - & - & - & 0.837 & - & - & - & - & - & 0.837 \\
    \textbf{NR} & gds142 & - & - & - & - & - & - & - & - & 0.826 &0.826 \\
    \textbf{NR} & LuisRamos07 & - & - & - & 0.822 & - & - & - & - & - & 0.822 \\
    \textbf{NR} & VerbaNexAI Lab \cite{morillo-EtAl:2024:SemEval2024} & - & - & - & 0.819 & - & - & - & - & - & 0.819 \\
    \textbf{NR} & Fired\_from\_NLP \cite{shanto-EtAl:2024:SemEval2024} & - & - & - & 0.810 & - & - & - & - & - & 0.810 \\
    \textbf{NR} & Roronoa\_Zoro & - & - & - & 0.810 & - & - & - & - & - & 0.810 \\
    \textbf{NR} & NLP\_STR\_teamS \cite{su-zhou:2024:SemEval2024} & - & - & - & 0.809 & - & - & - & - & - & 0.809 \\
    \textbf{NR} & DataJo & - & 0.356 & - & - & - & - & - & - & - & 0.356 \\
    \bottomrule
    \end{tabular}
    }
    \caption{Track A results. The best results are in bold, and NR stands for \textit{not ranked}. As the methods are highly language-dependent, we only rank teams that participated in at least 8 sub-tracks, but we highlight in blue the best results achieved by non-ranked teams. (Non-ranked teams are sorted based on the number of languages they participated in.)}
    \label{tab:track_a}
\end{table*}

\subsubsection{Best Performing Systems}
\paragraph{AAdaM}
They opted for data augmentation by translating the English SemRel dataset and STSB (semantic similarity) to create and augment data in other languages.
The team explored both fine-tuning and adapter-based tuning. Given a target language, they first fine-tuned the cross-encoder-based AfroXLMR model \cite{alabi-etal-2022-adapting} on the augmented data as a warm-up or TAPT (Task-Adaptive-Pre-Training) and then continued the fine-tuning on the provided SemRel data.

\paragraph{NRK}
They ensembled various BERT-like models and used a weighted voting technique to improve the performance of their model.

\paragraph{PEAR}
They examined the effect of combining or using per-language data through 5-fold validation. They did not conduct any text preprocessing to maintain fairness across languages. They defined three model configurations: ``base'' with no training, ``all'' trained on all languages, and ``lan'' trained on one language. They experimented with multilingual embeddings, cross-encoders, and augmented data from bi-encoders. 

\subsubsection{Popular Methods}
The general trend for the methods submitted to track A was (1)\ embedding sentence pairs into text and (2)\ training a regression model. Some teams used traditional embeddings and regression approaches (e.g., word2vec with support vector regressor -- team `Text Mining'). The majority used deep learning approaches (e.g., BERT, RoBERTa) or other large pre-trained transformer models (e.g., teams ``IITK'', ``Fired\_from\_NLP, HausaNLP''). When using these models, the teams would often experiment with different hyperparameters. Some teams went further and modified the specific learning approach or representations learned through methods such as contrastive learning (e.g., team: IITK). 

\subsubsection{Most Effective and Original Methods}
In track A, the participants used the provided training sets for each of the 9 languages included in the track (\texttt{amh, arq, ary, eng, esp, hau, kin, mar} and \texttt{tel}). Overall, the different teams explored several approaches to enhance the performance. For instance, the top performing team PALI, used MT-DNN (Multi-Task Deep Neural Networks for Natural Language Understanding) \cite{liu2019multi} and outperformed all the other teams across all languages except for Spanish and Kinyarwanda. For Kinyarwanda, king001 who used MT for data augmentation and multilingual mixed training and XLM-R \cite{conneau-etal-2020-unsupervised} as a base model achieved the best performance, and AAdaM who used translation-based data augmentation and adapter-based tuning reported the best score.

Note. however, that since PALI and king001 did not submit system description papers, they are not ranked in Tables \ref{tab:top_systems} and \ref{tab:track_a}.

\subsection{Task B: Unsupervised}
\begin{table*}[!ht]
    \centering
    \resizebox{\textwidth}{!}{%
    \begin{tabular}{rlccccccccccccc}
    \toprule
    \textbf{Rank} & \textbf{Team} & \textbf{\texttt{afr}} & \textbf{\texttt{amh}} & \textbf{\texttt{arb}} & \textbf{\texttt{arq}} & \textbf{\texttt{ary}} & \textbf{\texttt{eng}}& \textbf{\texttt{esp}} & \textbf{\texttt{hau}} & \textbf{\texttt{hin}} & \textbf{\texttt{ind}} & \textbf{\texttt{kin}} & \textbf{\texttt{pun}} & \textbf{\texttt{Average}} \\
    
    \midrule

    \textbf{1} & SATLab \cite{bestgen:2024:SemEval2024} & 0.761 & \textbf{0.764} & 0.487 & \textbf{0.521} & \textbf{0.599} & 0.774 & \textbf{0.709} & \textbf{0.513} & 0.649 & \textbf{0.491} & 0.458 & -0.215 & \textbf{0.543} \\
    \textbf{2} & MasonTigers \cite{goswami-EtAl:2024:SemEval2024} & 0.757 & 0.656 & 0.405 & 0.424 & 0.561 & 0.766 & 0.661 & 0.504 & 0.571 & 0.382 & \textbf{0.465} & 0.020 & 0.514 \\
    \textbf{3} & HW--TSC \cite{piao-EtAl:2024:SemEval2024} & 0.639 & 0.650 & 0.402 & 0.296 & 0.460 & 0.758 & 0.641 & 0.382 & 0.613 & 0.445 & 0.323 & \textbf{0.173} & 0.482 \\
    \textbf{4} & UAlberta \cite{shi-EtAl:2024:SemEval2024} & \textbf{0.789} & 0.723 & 0.467 & 0.368 & 0.063 & 0.775 & 0.680 & 0.380 & 0.691 & 0.484 & 0.378 & -0.027 & 0.481 \\
    \midrule
    \rowcolor{Gray}
    * & Lexical Overlap& 0.706 &	0.633 &	0.320 &	0.400 &	0.627 &	0.670 &	0.670 &	0.306 &	0.527 &	0.553 &	0.333 &	-0.274 &	0.456 \\ \midrule
    \textbf{5} & silp\_nlp \cite{singh-goyal-tiwary:2024:SemEval2024} & 0.732 & 0.643 & 0.314 & 0.402 & 0.552 & 0.317 & - & 0.387 & 0.571 & 0.532 & 0.350 & -0.110 & 0.400 \\
    \textbf{6} & HausaNLP \cite{salahudeen-EtAl:2024:SemEval2024} & 0.716 & 0.038 & 0.202 & 0.334 & 0.397 & 0.819 & 0.618 & 0.358 & 0.440 & 0.407 & 0.404 & -0.084 & 0.387 \\ \midrule
    \rowcolor{Gray}
    * & SemRel baseline (XLMR) & 0.562 & 0.573
& 0.316 & 0.247 & 0.174 & 0.601 & 0.689 & 0.041 & 0.507 &  0.467 & 0.132
 & -0.072  & 0.353 \\
      \midrule
    \textbf{NR} & IITK \cite{basak-EtAl:2024:SemEval2024} & - & 0.068 & - & 0.489 & 0.358 & 0.808 & 0.591 & 0.379 & - & - & - & - & 0.449 \\
    \textbf{NR} & YSP \cite{aali-hamidian-farinneya:2024:SemEval2024} & - & - & - & 0.385 & - & 0.788 & 0.598 & 0.193 & - & - & 0.377 & - & 0.468 \\
    \textbf{NR} & Tübingen-CL \cite{zhang-ltekin:2024:SemEval2024} & - & - & - & - & - & \best \textbf{0.837} & 0.705 & - & 0.649 & - & - & - & 0.730 \\
   \textbf{ NR} & CAILMD-23 \cite{sonavane-EtAl:2024:SemEval2024} & - & - & - & - & - & 0.819 & - & - & \best \textbf{0.797} & - & - & - & 0.808 \\
    \textbf{NR} & Self-StrAE \cite{opper-narayanaswamy:2024:SemEval2024} & 0.765 & - & - & - & - & - & 0.635 & - & - & - & - & - & 0.700 \\
   \textbf{NR} & NLU-STR \cite{malaysha-jarrar-khalilia:2024:SemEval2024} & - & - & \best \textbf{0.489} & - & - & - & - & - & - & - & - & - & 0.489 \\
    \bottomrule
    \end{tabular}
    }
    \caption{Track B results. The best results are in bold, and NR stands for \textit{not ranked}. As the methods are highly language-dependent, we only rank teams that participated in at least 8 sub-tracks, but we highlight in blue the best results achieved by non-ranked teams. (Non-ranked teams are sorted based on the number of languages they participated in.)}
    \label{tab:track_b}
\end{table*}

\subsubsection{Best Performing Systems}
\paragraph{SATLab}
Team SATLab used a system based on a model developed for authorship identification of source code \citep{bestgen-2019-cecl}. The system processed each pair of utterances independently, generating a distance between them without relying on additional information. Their pre-processing involved lower-casing of texts and making use of character $n$-grams ranging from 1 to 5 characters, encompassing all characters including spaces, punctuation marks, symbols, and characters from different writing systems. All $n$-grams were retained without a frequency threshold. The frequency of each feature was weighted by a logarithmic function, and the features of each statement were weighted by the L2 norm. The semantic similarity between utterances was estimated using the Euclidean distance between sets of $n$-grams in each pair. 

\paragraph{MasonTigers}

In the initial phase, team MasonTigers obtained the embeddings of training data instances and used TF--IDF, PPMI, LaBSE sentence transformer, and language-specific BERT models for multiple languages. Cosine similarity scores were then computed between pairs of embeddings, followed by the use of ElasticNet and Linear Regression separately to predict sentence pair similarity. Predicted values were clipped to ensure a range from 0 to 1.

\paragraph{HW--TSC}

Team HW-TSC's method included the $N$-gram chars utilising tokenizers from XLM-RoBERTa and m-BERT as key features to compute similarity scores based on $n$-gram dictionaries of sentences. They also used BERTScore to assess text quality based on the cosine similarity of token-level representations from the BERT model.

\subsubsection{Popular Methods}
As the main challenge with track B was the prevention of using any data of more than two words long related to semantics, many teams such as Hausa--NLP and Tübingen--CL used pre-trained language models such as All-MiniLM-L6-v2 \cite{reimers-2019-sentence-bert}. 

Most teams opted for language-specific data and models, if not trained on similarity data, and compared the performance to monolingual BERT models. However, none of these methods were used by the top three performing teams.

\subsubsection{Most effective and Original Methods}

The most effective methods for the unsupervised track for all languages were submitted by teams SATLab, MasonTigers, and HW--TSC (top--3). SATLab's approach involved processing pairs independently using character $n$-grams. MasonTigers, on the other hand, leveraged various embedding methods and statistical machine learning using simple features such as TF-IDF and BERT models to compute the cosine similarity between embeddings, further refined using ElasticNet. 
On the other hand, The HW--TSC team used innovative techniques such as the $N$-gram chars method with XLM-R and m-BERT tokenizers, as well as the BERTScore to evaluate the text quality. 

In Table \ref{tab:track_b}, we also have honorable mentions for teams that did not participate in all the languages but achieved remarkable results in one or a few languages. Notably, team CAILMD--23 achieved the best results in Hindi by using Hindi-BERT-v2, and team Tübingen--CL achieved the best results in English.  

\subsection{Task C: Crosslingual}
\begin{table*}[!ht]
    \centering
    \resizebox{\textwidth}{!}{%
    \begin{tabular}{rlccccccccccccc}
    \toprule
    \textbf{Rank} & \textbf{Team} & \textbf{\texttt{afr}} & \textbf{\texttt{amh}} & \textbf{\texttt{arb}} & \textbf{\texttt{arq}} & \textbf{\texttt{ary}} & \textbf{\texttt{eng}}& \textbf{\texttt{esp}} & \textbf{\texttt{hau}} & \textbf{\texttt{hin}} & \textbf{\texttt{ind}} & \textbf{\texttt{kin}} & \textbf{\texttt{pun}} & \textbf{\texttt{Average}} \\

    \midrule

        \textbf{1} & AAdaM \cite{zhang-EtAl:2024:SemEval20242} & 0.814 & 0.863 & 0.653 & 0.551 & 0.600 & 0.794 & 0.621 & 0.729 & 0.839 & \textbf{0.528} & 0.650 & \textbf{0.155} & \textbf{0.650} \\

    \textbf{2} & UAlberta \cite{shi-EtAl:2024:SemEval2024} & 0.806 & 0.816 & \textbf{0.671} & 0.441 & 0.602 & - & 0.572 & 0.678 & 0.828 & 0.449 & 0.636 & -0.017 & 0.589 \\
    \midrule
    \rowcolor{Gray}
    * & SemRel baseline (LaBSE) & 0.786 & 0.838 & 0.615 & 0.463 & 0.404 & 0.800 & 0.623 & 0.625 & 0.760 &  0.472 & 0.571 & -0.049 & 0.579 \\
          \midrule
      
    \textbf{3} & silp\_nlp \cite{singh-goyal-tiwary:2024:SemEval2024} & 0.747 & 0.805 & 0.427 & 0.387 & 0.673 & 0.737 & 0.569 & 0.643 & 0.801 & 0.472 & - & -0.037 & 0.566 \\
    \textbf{4} & MaiNLP \cite{zhou-EtAl:2024:SemEval2024} & 0.738 & 0.728 & 0.399 & 0.274 & 0.568 & - & - & - & 0.695 & 0.319 & \textbf{0.681} & 0.087 & 0.499 \\
    \textbf{5} & USTCCTSU \cite{li-EtAl:2024:SemEval20242} & 0.603 & 0.656 & 0.469 & 0.420 & 0.402 & 0.700 & 0.689 & 0.111 & 0.596 & 0.476 & 0.302 & -0.084 & 0.445 \\
    \textbf{6} & umbclu \cite{roydipta-vallurupalli:2024:SemEval2024} & \textbf{0.822} & 0.043 & 0.035 & 0.126 & -0.038 & 0.788 & 0.609 & 0.457 & 0.155 & 0.515 & 0.484 & -0.078 & 0.326 \\
    \textbf{7} & HausaNLP \cite{salahudeen-EtAl:2024:SemEval2024} & 0.737 & -0.031 & 0.184 & 0.074 & 0.276 & 0.360 & 0.604 & 0.177 & 0.346 & 0.472 & 0.319 & 0.114 & 0.303 \\
    \textbf{8} & MasonTigers \cite{goswami-EtAl:2024:SemEval2024} & 0.385 & 0.131 & 0.213 & 0.221 & 0.203 & 0.310 & 0.557 & 0.099 & 0.511 & 0.133 & 0.079 & 0.020 & 0.239 \\
    \textbf{NR} & USTC\_NLP & 0.749 & 0.709 & 0.517 & 0.414 & 0.613 & 0.784 & 0.685 & 0.476 & 0.658 & 0.460 & 0.454 & -0.248 & 0.523 \\
    \textbf{NR} & king001 & 0.810 & \best \textbf{0.878} & 0.657 & \best \textbf{0.614} & \best \textbf{0.820} & - & \best \textbf{0.708} & \best \textbf{0.733} & \best \textbf{0.844} & 0.376 & 0.630 & -0.050 & 0.641 \\

    \textbf{NR} & saturn & 0.818 & 0.814 & - & - & - & - & - & 0.569 & - & - & 0.604 & -0.103 & 0.540 \\
    \textbf{NR} & YSP \cite{aali-hamidian-farinneya:2024:SemEval2024} & - & - & - & 0.225 & - & \best \textbf{0.819} & 0.657 & 0.212 & - & - & 0.256 & - & 0.434 \\
    \textbf{NR} & CAILMD-23 \cite{sonavane-EtAl:2024:SemEval2024} & - & - & - & - & - & 0.786 & - & - & 0.810 & - & - & - & 0.798 \\
    \textbf{NR} & PALI & - & - & - & - & 0.842 & - & - & - & - & - & - & - & 0.842 \\
    \textbf{NR} & faridlazuarda & - & - & - & - & - & - & - & - & - & - & 0.600 & 0.058 & 0.329 \\
    \textbf{NR} & ETMS@IITKGP & - & - & - & - & - & - & 0.549 & - & - & - & - & - & 0.549 \\
    \textbf{NR} & Silp\_nlp & - & - & - & - & - & - & - & - & - & 0.472 & - & - & 0.472 \\
    \textbf{NR} & lukmanaj & - & - & - & - & - & - & - & 0.177 & - & - & - & - &  0.177 \\
    \bottomrule
    \end{tabular}
    }
    \caption{Track C results. The best results are in bold, and NR stands for \textit{not ranked}. As the methods are highly language-dependent, we only rank teams that participated in at least 8 sub-tracks, but we highlight in blue the best results achieved by non-ranked teams. (Non-ranked teams are sorted based on the number of languages they participated in.)}
    \label{tab:track_c}
\end{table*}

\subsubsection{Best Performing Systems}

\paragraph{AAdaM} They experimented with full fine-tuning, adapter fine-tuning using MAD \cite{pfeiffer-etal-2020-mad}, and data augmentation using different language combinations to augment data in a given source language.

\paragraph{UAlberta} They used an XGBoost regressor-based \citep{Chen_2016} ensemble approach to integrate the predicted relatedness scores of three distinct regression models, with one optional SBERT model, as input and returned the final relatedness score as output. %Each of these models uses a different pre-trained language model as its backbone, specifically RoBERTa Large, T5 Base, GPT-2 Base, and the optional SBERT (MPNet).
They applied the English version of their method trained for Track A to the translations of the non-English test sets.
The regression model fine-tuned on MPNet was used in the XGBoost ensemble only for \texttt{amh, hau,} and \texttt{hin}, but not for the other languages such  as \texttt{esp, ary, kin, ind, arb, arq,} and \texttt{afr}. The pre-trained English language models that were used include RoBERTa Large, T5 Base, and GPT2 Base, as well as MPNet only for languages \texttt{amh, hau,} and \texttt{hin}. 

\paragraph{silp\_nlp} They used the provided datasets and cross-lingual transferability with all the provided datasets, except data in the target language, as a source. Their cross-lingual transfer approach made use of MuRIL \citep{muril-lm} which led to the best results for Hindi and XLM-R \citep{conneau-etal-2020-unsupervised} led to the best ones for all the other languages.

\subsubsection{Popular Methods}
For the crosslingual track, many teams including best-performing ones such as UAlberta chose approaches similar to the ones used for supervised sub-tracks (e.g., using an XGBoost regressor \citep{Chen_2016} ). As the main challenge was to determine how to leverage data in languages other than the target, many teams combined the provided SemRel datasets in all possible languages (e.g., king001, AAdaM). 
Some used the training datasets without any modifications (e.g., team HausaNLP) and others experimented with different language combinations to use those that would lead to the best results (e.g., MasonTigers). 
Finally, some teams applied advanced techniques to modify the vector embedding space (e.g., by adjusting for the anisotropic nature of vector spaces -- team: USTCCTSU). 

\subsubsection{Most Effective and and Original Methods}
Overall, applying methods that are similar to the ones used in the supervised track using data in different languages can indeed lead to good results (e.g., king001, AAdaM, UAlberta). In addition, combining data in different languages and testing on another could boost the performance of crosslingual models for STR as shown by team sil\_nlp who achieved the best results in Amharic and Moroccan Arabic. 
Further, we note that leveraging advanced features such as (1)\ linguistic features (e.g., language family) as performed by MaiNLP, who achieved the best results for Kinyarwanda, and (2)\ embedding features by adjusting the distribution of the similarity scores as experimented by USTCCTSU could also help boost the performance.

Besides reporting on the best-performing teams only, in Table \ref{tab:track_c}, we also mention teams that did not participate in many sub-tracks but achieved good results such as team YSP, which outperforms all the other teams in English.

\section{Discussion}
We observe that in general, teams opt out of pre-trained models, and in most cases, the methods do not perform equally well across languages. Hence, for a given track, performing well in a language does not mean performing equally well in another language.

Further, the results show that good scores are not only related to low vs. high-resourcedness. For instance, In tracks B and C, results for Modern Standard Arabic (\texttt{arb}), which is considered high resource, are sometimes worse than those for low resource languages such as Amharic (\texttt{amh}) and Kinyarwanda (\texttt{kin}).

Interestingly, although the participating teams rarely use language-specific features, such approaches lead to good and interpretable results, as reported by e.g., team MaiNLP, who leveraged information about language families in Track C. 
We also note that for Track C, using a simple LaBSE baseline can achieve results that are better or comparable to more sophisticated techniques (see \citet{ousidhoum2024semrel2024} for language-specific baseline results).

\section{Conclusion}
We presented the first shared task on semantic relatedness, covering three tracks and 14 languages in total. 
The submitted systems were ranked based on the ranking of their predicted relatedness scores compared to the gold labels. 

We summarised the reported results, the best-performing methods, and the most effective, promising, and original ones. Overall, our findings on sentence representation techniques vary across the different languages and show that determining semantic textual relatedness is not a trivial task.

\section{Limitations}
As stated in \citet{ousidhoum2024semrel2024}, we acknowledge that there is no formal definition of what constitutes semantic relatedness and that our annotations may be subjective. To mitigate the issue, we share our guidelines and annotated instances so researchers in the community can expand on our work, replicate it, and study the disagreements in our data. We are also aware of the limited number of data sources and data variety in some low-resource languages involved. We do not claim that the datasets released represent all variations of these languages. However, they remain a good starting point as they were carefully picked, labeled, and processed by native speakers.

\section{Ethics Statement}

As stated in \citet{ousidhoum2024semrel2024}, we acknowledge all the possible socio-cultural biases that can come with our data due to the data sources or the annotation process. When building our datasets, we did avoid instances with inappropriate or offensive utterances, but we might have missed some.
Our goal was to identify common perceptions of semantic relatedness by native speakers, and our labels are not meant to be standardised for any given language as these are not fully representative of its usage.

\bibliography{anthology,custom,sys_desc,bib_semeval}
\bibliographystyle{acl_natbib}

\appendix

\section{Appendix: Track A--Best Performing Teams}
\paragraph{PALI and king001}

Both teams PALI and king001 did not submit a task description paper. king001 chose to use translation for data augmentation and multilingual mixed training. The team used XLM--R as their base model and DeBERTa--v3 \cite{he2021deberta}.

\paragraph{AAdaM}
Team AAdaM opted for translation-based data augmentation to increase the training data size for better performance. The English STR training data and STSB (semantic similarity) data were translated to create augmented datasets in other languages.
The team explored both fine-tuning and adapter-based tuning, aiming to examine and compare their effectiveness on STR across the different languages.
Given a target language, they first fine-tuned the cross-encoder-based AfroXLMR model on the augmented data as a warm-up or TAPT (Task-Adaptive-Pre-Training) and then continued the fine-tuning on the provided STR data.

\paragraph{NRK}
They used ensembling and various BERT-like models.

\paragraph{PEAR}
They examined the effect of combining vs. using language-specific data through 5-fold validation. No text preprocessing was conducted to maintain fairness across languages. Three model configurations were defined: ``base'' with no training, ``all'' trained on all languages, and ``lang'' trained on one language. They experimented with multilingual embeddings, cross-encoders, and data augmentation with bi-encoders. Parameter optimization was conducted using Optuna. 

\paragraph{silp\_nlp} 
Team silp\_nlp's methodology for track A was a two-stage training.
In the initial stage, they trained a model using all 9 languages covered in track A with MuRIL \citep{muril-lm}. They experimented with different hyperparameters on five epochs and selected the best multilingual checkpoint based on the average validation data loss. They fine-tuned the resulting model using the training data for each language in track A and ended up with monolingual models. 

Each monolingual model was trained using different hyperparameters and they selected their final model based on the validation data loss of the corresponding language track.

\paragraph{NLP\_1@SSN}
They used SBERT fine-tuned on multilingual and monolingual pre-trained language models Overall, they observed that the usage of monolingual PLMs did not guarantee better performance.

\paragraph{UAlberta} They used an ensemble approach with an XGBoost regression \citep{Chen_2016} to integrate the predicted relatedness scores of three distinct regression models, with one optional SBERT model, as input and returned the final relatedness scores as output. Each of these models used a different pre-trained language model as its backbone, specifically RoBERTa Large \citep{liu2019roberta}, T5 Base, GPT-2 Base, and the optional SBERT (MPNet).
They merged the English training and development sets with the translated training set of the target language. 
%The translated development set of the target language is kept as it is.
%They also translated the English data to esp, hau, amh, mar, tel and used it for each language, respectively. 
Then, they split them again via uniform random sampling according to their original sizes to establish new training and development splits. The did not use the data provided for arq, ary, and kin, and applied the English-trained version of their method to the English translations of the arq, ary, and kin test sets instead. 

\paragraph{MBZUAI-UNAM}
They fine-tuned a paraphrase model architecture to train language-specific models, using a separate pre-trained model to embed each language. They also experimented with combined training sets based on the language families. 

\paragraph{INGEOTEC}
For English and Spanish, they used embeddings (microsoft/mpnet-base, bert-base-multilingual-cased) to train an SVM classifier.
For the other languages, they used prior work EvoMSA.

\paragraph{HausaNLP}
They used different base pre-trained models.

\section{Appendix: Track B}
\paragraph{SATLab}
They proposed a system based on a model developed for the authorship identification of source code \citep{bestgen-2019-cecl}. It processed each pair of utterances independently, generating a distance between them without relying on additional information. Pre-processing involved lower-casing of texts. Character $n$-grams ranging from 1 to 5 characters are used, encompassing all characters including spaces, punctuation marks, symbols, and characters from different writing systems, all $n$-grams are retained without a frequency threshold. The frequency of each feature was weighted by a logarithmic function, and the features of each statement were weighted by the L2 norm. Semantic similarity between utterances was estimated using Euclidean distance between sets of $n$-grams in each pair. 

\paragraph{MasonTigers}

In the initial phase, team MasonTigers obtained embeddings of training data and used various methods including TF-IDF, PPMI, LaBSE sentence transformer, and language-specific BERT models for multiple languages. Cosine similarity was then computed between pairs of embeddings, followed by applying ElasticNet and Linear Regression separately to predict sentence pair similarity in the development phase. Predicted values were clipped to ensure a range from 0 to 1.

\paragraph{HW--TSC}

The key features used by team HW-TSC's method included the $N$-gram chars method using XLM-RoBERTa and m-BERT tokenizers to compute similarity scores based on $n$-gram sentence dictionaries. They also used the BERTScore method to assess text quality based on the cosine similarity of token-level representations from the BERT model. 

\paragraph{UAlberta}
They used a linear combination of two sets of normalized results, each derived from the cosine similarity measurements of sentence embeddings obtained from the hidden sentence representations processed by BERT Large and RoBERTa Large.
They calculated the final relatedness scores by averaging the cosine similarity scores of sentence embeddings obtained from each set.

\paragraph{silp\_nlp}
They converted the sentences into unigram and bigram representations and used Support Vector Regression (SVR). 

Sentences were combined and transformed into a vector, and each sentence was indexed based on a value that represented the count of unigrams/bigrams present in it. The resulting vector was fed into the SVR model along with label values for training.

\paragraph{HausaNLP}
Team HausaNLP used a standard all-MiniLM-L6-v2 model to train a model for Track B.

\paragraph{IITK}
Team IITK uses SimCSE \citep{gao-etal-2021-simcse}, or Simple Contrastive Learning of Sentence Embeddings that induced slight variations in its representation through dropout.
TSDAE\citep{wang-etal-2021-tsdae-using}, a denoising autoencoder, was used to generate sentence embeddings by reconstructing original sentences in the presence of noise. They used BERT to construct the denoising autoencoder and TSDAE optimized the likelihood of reconstructing sentences during training, which led to compact embeddings.

\paragraph{Tübingen-CL}
Team Tübingen-CL opted for exploring features like cosine distance of average word embeddings and word overlap ratios, to potentially enhance performance. For English, they used two models: multi-qa-MiniLM-L6-cos-v1 trained on QA pairs and trained for semantic search and e5-base-unsupervised trained on various pairs including question-answer and post-comment pairs, both refined with unsupervised transformation (PCA). Two additional features, PCA-transformed GloVe embeddings, and content word overlap ratios were incorporated into the unsupervised ensemble system. Similar methods were applied for Spanish and Hind using multilingual BERT embeddings and various feature combinations to predict relatedness.

\paragraph{CAILMD-23}

Team CAILMD-23 participated in the English and Hindi sub-tracks of the unsupervised task. They experimented with a few models such as BERT-based and Hindi-Bert v2. The latter is trained on Hindi text comprehension with a training corpus of roughly 1.8 billion tokens.

\section{Appendix: Track C}
\paragraph{AAdaM}
hey experimented with full fine-tuning, adapter fine-tuning using MAD \cite{pfeiffer-etal-2020-mad}, and data augmentation using different language combinations to augment data in a given source language.

\paragraph{king001}
They did not submit a system description paper but they reported combining the training datasets provided for track A, and if one of them was in the target language, they translated it into English. Then, they run multi-task learning for 15 epochs.

\paragraph{UAlberta} They used an ensemble approach with an XGBoost regressor \citep{Chen_2016} to integrate the predicted relatedness scores of three distinct regression models, with one optional SBERT model, as input. Each of their models used a different pre-trained language model as its backbone, specifically RoBERTa Large, T5 Base, GPT-2 Base, and the optional SBERT (MPNet).

They applied the English version of their method reported for Track A to the translations of the non-English test sets.
The regression model fine-tuned on MPNet was used in the XGBoost ensembling method for amh, hau, and hin and not for esp, ary, kin, ind, arb, arq, and afr. 

\paragraph{silp\_nlp} They used cross-lingual transferability on all the provided datasets except for the target language (e.g., when they test on Telugu, they use all languages except Telugu). In their cross-lingual transfer approach, MuRIL \citep{muril-lm} led to the best results for Hindi and XLM-R \citep{conneau-etal-2020-unsupervised} for all the other languages.

\paragraph{USTCCTSU} They used XLM-R \citep{conneau-etal-2020-unsupervised} trained on a combination of language inputs (chosen by trying different combinations with the best one including all the languages). They ranked in the top 5  for ind,arq, and esp. 

They adjusted the similarity scores for the XLM-R base models by applying a technique called \textit{whitening} that allowed them to change the non-uniform score distribution into multiple distributions, and eventually, into a uniform one.

\paragraph{MaiNLP} They finetuned multilingual LLMs (XLM-R and Furina) using an upscaled version of the data from Track A. They assessed the linguistic similarity of the available Track A data to determine the most useful datasets and experimented with different language families.
For pre-processing, they used tokenization, segmentation, and translation. They also experimented with transliteration to change the scripts into Latin. Translations helped them upscale the English, Hausa, and Spanish training data and then evaluate on the Track C data. They achieved the best results for Kinyarwanda.

\paragraph{umbclu}
They pre-trained T5 models with SemRel data. They used the English fine-tuned models for inference on all language test sets except English. On the other hand, they used Spanish models for inference on English.

\paragraph{HausaNLP}
They used a BERT-based model fine-tuned on the datasets in other languages. E.g., they trained on English data and tested on Spanish, trained on Kinyarwanda and tested on Hausa. 
They ranked in the top 5 in Task C  for ind, pan.

\paragraph{MasonTigers}
They used statistical machine learning (Linear Regression, ElasticNet with TF-IDF and PPMI features) along with language-specific BERT-based models to predict the relatedness scores. The models were trained on dataset combinations of 5 languages other than the target language and used BERT-based models's similarity prediction on the target test data (e.g., they trained on amh, eng, esp, arq, ary and tested on afr).
For language-specific BERT-like models, they used African language BERT-base models, Arabic BERT-based models, AfricanBERTa, and for eng, hin, ind, pun, esp, they used spanBERTa, BanglaBERT, RoBERTa-tagalog-base-BERT, HindiBERT, and RoBERTa.

\begin{table*}[]
    \centering
    \begin{tabular}{lc}
    \toprule
         \textbf{Team}&\textbf{Paper}  \\
         \midrule
        AAdaM&\citet{zhang-EtAl:2024:SemEval20242}\\
        All--Mpnet&\citet{siino:2024:SemEval20248}\\        
        BITS Pilani&\citet{venkatesh-raman:2024:SemEval2024}\\
        CAILMD--23&\citet{sonavane-EtAl:2024:SemEval2024}\\
        Fired\_from\_NLP&\citet{shanto-EtAl:2024:SemEval2024}\\
        HausaNLP &\citet{salahudeen-EtAl:2024:SemEval2024}\\         
        HW-TSC&\citet{piao-EtAl:2024:SemEval2024}\\
        IITK&\citet{basak-EtAl:2024:SemEval2024}\\
        INGEOTEC&\citet{moctezuma-tellez-graff:2024:SemEval2024}\\            
        %LinguAlchemy &\citet{Adilazuarda_semrel2024}\\
        MaiNLP&\citet{zhou-EtAl:2024:SemEval2024}\\
        MasonTigers&\citet{goswami-EtAl:2024:SemEval2024}\\        
        MBZUAI--UNAM&\citet{ortizbarajas-belenguix-gomzadorno:2024:SemEval2024}\\
        NLP--LISAC &\citet{benlahbib-EtAl:2024:SemEval2024}\\
        NLP\_STR\_teamS&\citet{su-zhou:2024:SemEval2024}\\
        NLP\_Team1\@SSN&\citet{b-EtAl:2024:SemEval2024}\\
        NLU--STR&\citet{malaysha-jarrar-khalilia:2024:SemEval2024}\\
        NRK &\citet{nguyen-thin:2024:SemEval2024}\\
        OZemi &\citet{takahashi-EtAl:2024:SemEval2024}\\
        PEAR&\citet{jrgensen:2024:SemEval2024}\\
        Pinealai&\citet{eponon-ramosperez:2024:SemEval2024}\\
        SATLab &\citet{bestgen:2024:SemEval2024}\\
        scaLAR&\citet{m-m:2024:SemEval2024}\\
        Self--StrAE &\citet{opper-narayanaswamy:2024:SemEval2024}\\        
        SemanticCUETSync&\citet{hossain-EtAl:2024:SemEval2024}\\
        Sharif\_STR&\citet{ebrahimi-EtAl:2024:SemEval20242}\\     silp\_nlp&\citet{singh-goyal-tiwary:2024:SemEval2024}\\        TECHSSN&\citet{g-EtAl:2024:SemEval2024}\\        
        Text Mining&\citet{keinan:2024:SemEval2024}\\
        Tübingen--CL&\citet{zhang-ltekin:2024:SemEval2024}\\        UAlberta&\citet{shi-EtAl:2024:SemEval2024}\\        UMBCLU&\citet{roydipta-vallurupalli:2024:SemEval2024}\\
        USTCCTSU&\citet{li-EtAl:2024:SemEval20242}\\
        VerbaNexAI&\citet{morillo-EtAl:2024:SemEval2024}\\
        WarwickNLP&\citet{ebrahim-joy:2024:SemEval2024}\\
        YNU--HPCC&\citet{li-wang-zhang:2024:SemEval2024}\\        YSP&\citet{aali-hamidian-farinneya:2024:SemEval2024}\\
        \bottomrule
    \end{tabular}
    \caption{The participating teams (alphabetically ordered) that submitted system description papers.}
    \label{tab:sys_desc}
\end{table*}

\end{document}